\documentclass{article}

\usepackage[final]{cpal_2026}

\usepackage{microtype}
\usepackage{graphicx}
\usepackage{amsmath, amsthm}
\usepackage{booktabs}
\usepackage{url}
\usepackage{hyperref}
\usepackage{xcolor}
\usepackage{wrapfig}
\usepackage{verbatim}
\usepackage{amsfonts}

\newcommand{\x}{\mathbf{x}}

\newcommand{\bu}{\mathbf{u}}
\newcommand{\bv}{\mathbf{v}}

\newcommand{\bs}{\mathbf{s}}
\newcommand{\bP}{\mathbf{P}}

\newtheorem{lemma}{Lemma}

\title{Prompt Stability Matters: Evaluating and Optimizing Auto-Generated Prompt in General-Purpose Systems}

\author{%
  Ke Chen\textsuperscript{1},
  Xucheng Yu\textsuperscript{1},
  Yufei Zhou\textsuperscript{2},
  Haohan Wang\textsuperscript{1} \\
  \textsuperscript{1}School of Information Sciences, University of Illinois Urbana-Champaign, USA \\
  \textsuperscript{2}Department of Economics, Duke University, USA \\
  \texttt{kec10@illinois.edu, xy63@illinois.edu, yz597@duke.edu, haohanw@illinois.edu}
}

\begin{document}

\maketitle

\begin{abstract}
Automatic prompt generation plays a crucial role in enabling general-purpose multi-agent systems to perform diverse tasks autonomously. Existing methods typically evaluate prompts based on their immediate task performance, overlooking the intrinsic qualities that determine their reliability. This outcome-centric view not only limits interpretability but also fails to account for the inherent stochasticity of large language models (LLMs). In this work, we bring attention to prompt stability—the consistency of model responses across repeated executions—as a key factor for building robust and effective prompt generation systems. To quantify this, we propose semantic stability as a criterion for assessing the response consistency of prompts. Based on the proposed metric, we developed the first stability-aware general-purpose prompt generation system that leverages stability feedback to iteratively enhance both prompt quality and system-level performance. Furthermore, we establish a logical chain between prompt stability and task success by analyzing the structural dependencies within our system, proving stability as a necessary condition for effective system-level execution. Empirical results across general and domain-specific tasks demonstrate that our stability-aware framework improves both accuracy and output consistency. By shifting the focus from one-off results to persistent reliability, our work offers a new perspective on prompt design and contributes practical tools for building more trustworthy general-purpose systems. 
The UI and source code for our system are publicly available at: \url{https://xucheng63.github.io/MyDataPilot/}.
\end{abstract}

\section{Introduction}

Imagine a project manager acting as a planner. Their job is to decompose a complex project into individual tasks, assign them to team members, and write clear instructions for each one. The project’s success hinges on the clarity of these instructions—any ambiguity may lead to misinterpretation, misalignment, and ultimately, failure to realize the original vision.

Now, imagine the same scenario with a twist: both the project manager and the team are highly versatile, capable of handling tasks across diverse domains. This general-purpose setting magnifies the challenge, as domain-specific language increases the risk of misunderstanding. In such context, instructions must be precise, consistent, and unambiguous.

This is no longer hypothetical. While human teams rarely have such general-purpose abilities, AI agents increasingly do. General-purpose multi-agent systems built on large language models (LLMs) assign roles and responsibilities via prompts \cite{shen2023hugginggpt, li2023camelcommunicativeagentsmind, wu2024autogen, Park2023GenerativeAgents, liu2025foundation,chen2025llm_ds_agent}, which define agent behavior, interaction, and decision-making. When prompts are ambiguous or unstable, agents may misinterpret roles, leading to coordination breakdown and unpredictable system behavior.

This fragility is worsened by the stochasticity of LLMs \cite{atil2024,jinevaluating,zhang2025achilles}. A prompt that yields a reasonable response once may generate a different one under the same conditions \cite{wang2023selfconsistency, moore2024, ribeiro-etal-2020-beyond}. In multi-agent systems where agents depend on one another’s outputs, such inconsistencies can cascade. Thus, we argue that prompts should be evaluated not only for correctness or accuracy, but also for \emph{stability}—their ability to consistently elicit semantically coherent responses across executions.

One of the key factors that determines whether general-purpose systems can achieve robust and reliable performance is prompt construction. Task-oriented systems typically rely on human-written prompts tailored to specific tasks, ensuring clarity and alignment \cite{llmfewshot, mishra2022crosstask, schick2021exploitingclozequestionsshot, wei2023chainofthoughtpromptingelicitsreasoning}. In contrast, general-purpose systems often depend on automatic prompt generation mechanisms, which offer scalability but may produce prompts that are poorly aligned with task or model behavior. Most existing methods for automatic prompt generation adopt an outcome-driven perspective, evaluating prompts based on downstream performance metrics such as task success rate or accuracy \cite{autoprompt, li2021prefixtuning, deng2022rlprompt, zhou2023largelanguagemodel}. While this provides a practical measure of effectiveness, it essentially evaluates the result of the prompt rather than the prompt itself. A particularly critical yet often overlooked limitation of this approach is its inability to account for \emph{prompt stability}, the consistency of a prompt’s outputs across repeated executions.

In stochastic LLMs, a prompt that succeeds once may fail when repeated, as minor changes in phrasing, sampling, or surrounding context can alter the model’s output. This lack of reproducibility poses a fundamental challenge for applications that require stable and predictable responses, where occasional success is insufficient for reliable operation.

However, despite its importance, prompt stability remains underexplored, in part due to the lack of frameworks to define or measure it \cite{liu2021}. Unlike performance, which can be evaluated per run, stability requires reasoning over distributions of outputs. Moreover, prompts are usually treated as static inputs, overlooking the output variance introduced by LLM sampling dynamics.

This paper presents \emph{prompt stability} as a core design objective in automated prompt generation. We argue that prompt quality should be judged not only by outcome success, but also by the ability to achieve it consistently under varying conditions. 

To justify this principle, we first conduct a theoretical analysis that formally links prompt stability to system-level execution quality. Through structural modeling and probabilistic reasoning, we show that variability in prompt interpretation can lead to significant deviations from the planner’s intended output, particularly in multi-agent workflows. We further conduct empirical validation to prove that compared with traditional metrics (e.g., mutual information~\cite{sorensen2022mintip}, prompt entropy~\cite{lu-etal-2022-fantastically}, clarity~\cite{ghost2025promptqual}, etc.), our proposed metric has better interpretability of prompt success rate.

Building on this insight, we introduce a unified framework for evaluating and optimizing prompt stability in general-purpose LLM-based systems. We propose \emph{semantic stability}, a novel metric that quantifies robustness by measuring the consistency of LLM outputs across repeated executions. Based on this metric, we design a self-optimizing prompt generation system, Promptor, that leverages stability feedback to iteratively refine prompt quality.

Through formal reasoning and empirical studies, we demonstrate that improving prompt stability leads to higher accuracy, reduced output variance, and greater reliability across a range of tasks and domains—including applications in finance, biology, and chemistry.

\section{Related Work}


\subsection{Performance-driven prompt optimization in multi-agent systems} 

Recent studies have explored automated prompt optimization to improve general-purpose language model systems. \cite{c1} propose LLM-AutoDiff, a framework inspired by automatic differentiation that updates prompts using natural language feedback derived from task performance. The system generates textual explanations of errors and uses another LLM to revise prompts accordingly. While it supports fine-grained tuning within multi-component workflows and incorporates structural cues such as prompt composition and revision history, its optimization remains fundamentally output-driven—guided by task-level metrics such as accuracy or factual consistency. It does not explicitly assess intrinsic prompt properties such as semantic stability or output variance across repeated executions.

Beyond single-round feedback-based refinement, more recent works have introduced iterative and closed-loop optimization strategies. REVOLVE \cite{zhang2024revolve} tracks how model responses evolve across multiple rounds of textual optimization. Instead of relying solely on feedback from the current iteration, it analyzes the trajectory of response changes over time, identifying whether the model exhibits stable improvement, stagnation, or oscillation. By incorporating response evolution trends into the refinement process, REVOLVE simulates a second-order optimization effect, leading to more stable and directed updates that help escape local optima in reasoning, problem-solving, and code generation tasks. 

Similarly, SIPDO \cite{yu2025sipdo} adopts a closed-loop optimization framework via synthetic data feedback. Rather than passively evaluating prompts on fixed datasets, SIPDO enables the model to generate increasingly challenging test cases to probe the weaknesses of the current prompt. Through a self-generated cycle of question construction, error identification, and prompt revision, the system progressively strengthens the prompt until performance stabilizes. This “generate–diagnose–refine” loop enhances robustness under diverse task scenarios.

Although these approaches advance prompt optimization beyond static evaluation, they remain primarily performance-driven. Their objective functions are defined in terms of downstream task success or error correction. As a result, they focus on improving observable outcomes, without explicitly modeling the stochastic variability of model responses under repeated executions. In contrast, our work shifts the focus from outcome improvement alone to intrinsic prompt stability, introducing semantic stability as a principled criterion for evaluating and optimizing consistency in general-purpose systems.



\subsection{Existing prompt evaluation methods}

Several works have focused on evaluating prompt quality from different angles. \cite{c2} introduce Automatic Prompt Engineer (APE), which scores prompts by the log-likelihood of model outputs, using it as a proxy for model confidence. 

\cite{c3} propose PromptEval, which assesses prompt effectiveness across examples via a probabilistic model based on item response theory (IRT). This framework estimates the likelihood that a prompt yields correct outputs, capturing both average performance and variability. However, it does not evaluate a prompt’s stability when repeatedly applied to the same input.

These approaches contribute valuable tools for evaluating prompts in terms of accuracy, confidence, and expected performance. However, they do not consider whether a prompt can elicit semantically consistent outputs across repeated runs—an important property for building reliable systems. In other words, the stability of a prompt’s behavior under varying model conditions remains largely unexamined.

\section{Motivations and Analytical Results}

We consider a planner--executor structure to analyze how prompt stability affects the overall behavior of a general-purpose multi-agent system. Let $P$ denote the planner module, which takes a task description $t$ (e.g., ``Given a large set of scRNA-seq data from multiple tissue types, identify major cell subpopulations, infer their lineage relationships, and generate hypotheses about potential marker genes for each subpopulation'') and decomposes it into a set of task-specific prompts. This process can be written as:
\[
\begin{aligned}
P(t) &\mapsto \{p_1, p_2, \dots, p_n\}, \qquad
\x_i = A_i(p_i).
\end{aligned}
\]

where $p_i$ denotes the $i$\textsuperscript{th} prompt, including both textual instructions and assigned data, to be sent to executor agent $A_i$. Note that since $A_i$ is a generic agent, its output can be of various formats either in texts, codes, or processed data. Without loss of generality, we can find certain encoding schema to encode the output into a scalar. Therefore, we collect these outputs into a diagonal matrix:
\[
\mathbf{X} = \mathrm{diag}(\x_1, \x_2, \dots, \x_n),
\]
and define the final output of the multi-agent system as:
\[
\bs = \bu^T \mathbf{X} \bv
\]
where $\bu, \bv \in \mathbb{R}^n$ encode aggregation vectors across agents.

Importantly, given a fixed task assignment, all randomness in the system originates from the LLMs (planner or executors). We use $\Lambda$ to denote the conditional distribution governing this stochastic behavior. For each executor $A_i$, we denote its corresponding stochastic process by $\Lambda^{(i)}$, which reflects the variability introduced by the LLM’s autoregressive sampling. Therefore, we have:
$\x_i \sim \Lambda^{(i)}(p_i)$.

We conjecture that one of the key challenges of a general-purpose multi-agent system is the ambiguity induced by the prompts $p_i$ constructed by the planner. When the planner $P$ writes the prompts $p_i$, these prompts are supposed to guide the LLM to complete the task in a way that aligns with the planner's own interpretation. In other words, if we use $\Lambda^\star$ to interpret these prompts, we obtain:
\[
\hat{\x_i} = \Lambda^\star(p_i), \quad
\hat{\bs} = \bu^T \hat{\mathbf{X}} \bv  =  \sum_{i=1}^n u_i v_i \hat{\x_i},
\]
where $\hat{\bs}$ represents the ideal system behavior as envisioned by the planner.

However, these prompts are not interpreted by the planner itself, but by executor agents using their own sampling processes (i.e., $\Lambda^{(i)}$). Even if the models are parameterized identically, stochastic sampling may yield diverging interpretations. Therefore, the actual execution gives:
\[
\x_i = \Lambda^{(i)}(p_i), \quad \bs = \sum_{i=1}^n u_i v_i \x_i.
\]
The deviation between the envisioned and actual outputs is:
\[
\left| \bs - \hat{\bs} \right| =  \left| \bu^T (\mathbf{X} - \hat{\mathbf{X}}) \bv \right| = \left| \sum_{i=1}^n u_i v_i (\x_i - \hat{\x_i}) \right|.
\]

\begin{lemma}
    With assumptions that $\x_i$ are independent, we have
    \[
\bP\left( \left| \bs - \hat{\bs} \right| \geq \epsilon \right) \leq 2 \exp\left( \frac{-\epsilon^2}{2 \sum_{i=1}^n (u_i v_i)^2 \text{Var}(\x_i) } \right).
\]
\end{lemma}
The proof is in Appendix \ref{proof}.

\textbf{Conclusion.}
This analysis shows that the deviation between actual and intended system outputs is primarily driven by the variance of executor responses. Even with clear prompts, LLM stochasticity can amplify small inconsistencies across agents. Reducing this variance by stabilizing the prompt–response relationship offers a principled path to improving system reliability—making prompt stability not just a heuristic goal, but a theoretically grounded optimization target.

\section{Method}

\subsection{Evaluate prompt via semantic stability}

As discussed in Section~3, the deviation between the system’s actual and intended outputs is closely tied to the variance of individual agent responses $\text{Var}(\x_i)$. However, since $\x_i$ is typically a generated string, direct variance computation is impractical. We thus seek a proxy that captures the consistency of outputs generated from the same prompt.

A natural idea is to model the token-level distributions and measure their divergence using KL divergence \cite{kl}. However, such methods overlook semantic differences—for instance, “I agree with the statement.” and “I do not agree with the statement” may share similar token distributions but convey opposite meanings. This motivates embedding-based evaluation \cite{Zhang2020BERTScore:, reimers-gurevych-2019}.

Specifically, we encode each sampled output $y_i$ into a semantic vector $v_i = \phi(y_i)$ using a pre-trained embedding model. In high-dimensional space, Euclidean distance becomes unreliable due to the \emph{curse of dimensionality}, so we use cosine distance, which is more robust for natural language applications \cite{cer-2018, reimers-gurevych-2019}.

For any output pair, the cosine distance is computed as:
\[
d_{ij} = 1 - \frac{v_i \cdot v_j}{\|v_i\| \|v_j\|}, \quad v_i = \phi(y_i)
\]

We define the \textit{semantic stability} $S(p)$ of a prompt $p$ as the average pairwise cosine similarity:
\[
S(p) = 1 - \frac{2}{N(N-1)} \sum_{i < j} d_{ij}
\]

Higher values of $S(p)$ indicate greater semantic consistency and thus stronger prompt stability, serving as a practical proxy for output variance in stochastic LLM settings.

\subsection{Stability-guided optimization framework}
Section~3 shows that the deviation between the planner’s expected output $\hat{\bs}$ and the actual system output $\bs$ is governed by executor variance. Ultimately, what matters is alignment with the true task goal $\bs^\star$, which is typically unobservable. 
Since $\bs^\star$ cannot be directly accessed, we instead introduce a proxy target $\bs^{\star'}$, 
representing the best output that a high-performing LLM can reasonably achieve within its capability. 
Optimizing toward $\bs^{\star'}$ therefore serves as the closest operational objective available to the system.
This yields the following decomposition with respect to the proxy target:
\[
|\bs^{\star'} - \bs| \leq |\bs^{\star'} - \hat{\bs}| + |\hat{\bs} - \bs|
\]

Accordingly, our system jointly optimizes two objectives:

\begin{align} \textbf{(1) Stability Objective:} \ & \max_{p_i} \; S(p_i) \ \  \textbf{(2) Planner Alignment Objective:} \ \min_{p_i} \; |\bs^{\star'} - \hat{\bs}| \notag \end{align}

\begin{figure}[h]
\centering
\includegraphics[width=\linewidth]{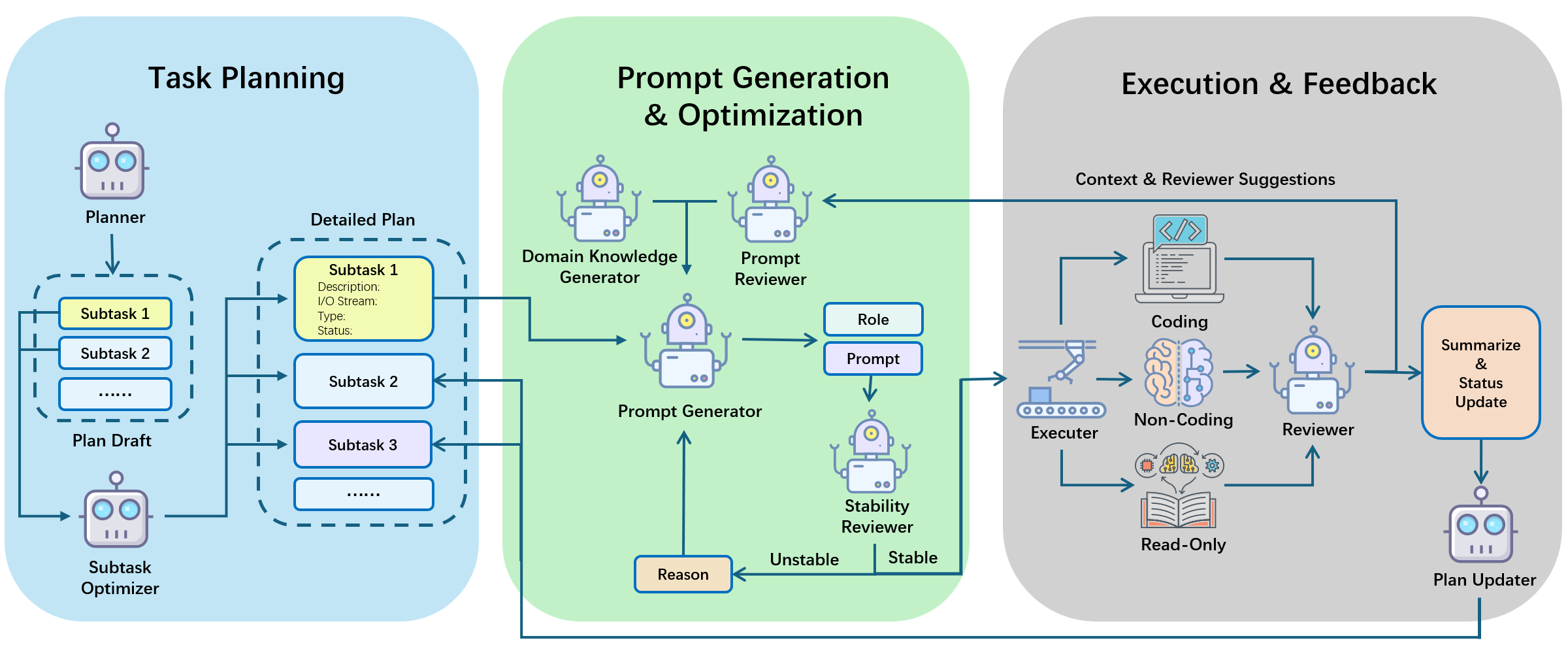}
\caption{Promptor system pipeline of our stability-aware prompt generation framework.}
\label{fig:system_pipeline}
\end{figure}

The first reduces execution variance via stability-aware prompt generation; the second improves task decomposition by minimizing the gap between planner intent and idealized output. Figure~\ref{fig:system_pipeline} illustrates how these objectives are operationalized in our system.

\subsubsection{Reducing Execution Deviation.}
To reduce $\text{Var}(\x_i)$, we directly optimize the semantic stability $S(p_i)$ for each prompt:
\[
\text{Var}(\x_i) \propto 1 - S(p_i)
\]
Prompts with $S(p_i) < \tau$ (a predefined threshold) trigger a Reviewer Agent, which diagnoses instability and revises the responsible component.

Each prompt $p_i$ is modularized as:
\[
p_i = [r_i, q_i, k_i, h_i]
\]
with $r_i$: role definition, $q_i$: task requirements, $k_i$: domain knowledge, and $h_i$: context/history. The Reviewer identifies the unstable subcomponent $z \in \{r_i, q_i, k_i, h_i\}$ and revises it to $z'$, forming an updated prompt $p_i'$. This refinement continues until
$S(p_i^{(t)}) \geq \tau$.

Once a prompt is deemed stable, a Summarizer Agent distills its raw output $\bs$ into a structured summary $\hat{\bs}$. The summarizer has access to the global task description and system plan, enabling it to selectively preserve only the information most relevant to the planner's intent. This mechanism helps align the actual output with the planner's expected structure, effectively reducing the execution deviation $|\bs - \hat{\bs}|$.

\subsubsection{Reducing Planner Error.}
To reduce $|\bs^{\star'} - \hat{\bs}|$, we implement mechanisms that refine both the planner's initial decomposition and its dynamic updates.

\textbf{Subtask Optimization.}  
Each subtask $t_i$ inherently has an ideal granularity $g_i^\star$: if $t_i$ is too coarse, it becomes ambiguous or overloaded with multi-step goals; if too fine, it may cause contextual fragmentation and hinder global coordination. Let $g(t_i)$ denote the actual granularity. Then,
\[
\left| \bs^{\star'} - \hat{\bs} \right| \propto \sum_{i=1}^n \left| g(t_i) - g_i^\star \right|
\]
Our Subtask Optimizer minimizes this quantity by adjusting subtask boundaries, types, and I/O specifications, thereby improving the alignment between the planner's internal objectives and the ideal output.

\textbf{Plan Updater.}  
To further reduce the planner-side deviation $|\bs^{\star'} - \hat{\bs}|$, we employ a Plan Updater that iteratively adjusts future subtasks based on the observed results of completed ones. Let $\hat{\bs}^{(t)}$ represent the planner’s intermediate output up to step $t$, and $\mathcal{T}_{>t}$ denote the set of unexecuted subtasks following $t$. The updater injects a corrective signal $\Delta_t$ into each $t_j \in \mathcal{T}_{>t}$:
\[
\hat{\bs}^{(t+1)} = \hat{\bs}^{(t)} + \sum_{j>t} \Delta_t^{(j)}
\]
where each $\Delta_t^{(j)}$ modifies the formulation or structure of future subtask $t_j$ in response to execution feedback at step $t$.

The form of $\Delta_t^{(j)}$ depends on whether subtask $t$ succeeds or fails.  
If $t$ succeeds, $\Delta_t^{(j)}$ propagates structural patterns, technical details, or constraints from $\x_t$ to improve the setup of $t_j$, enhancing clarity and knowledge continuity.  
If $t$ fails, it triggers a strategy shift—reformulating or replacing $t_j$, adjusting assumptions, or invoking fallback plans—guiding $\hat{\bs}$ toward a more achievable trajectory.

In both cases, the Plan Updater integrates feedback into future subtasks, refining the planner’s trajectory $\hat{\bs}$ and reducing accumulated misalignment. This iterative process helps minimize the gap between planner intent and the ideal output:
$|\bs^{\star'} - \hat{\bs}| \to \min$

\subsection{Engineering Specifications}

While our system is theoretically grounded in the planner–executor structure and stability-aware prompt optimization framework, its practical success also depends on several engineering components beyond the core method. Modules like the \textit{Domain Knowledge Generator}, \textit{Executor Module}, and other agents shown in Figure~\ref{fig:system_pipeline} are essential for robust performance. 
In addition, we designed a lightweight UI to support interactive usage and facilitate seamless integration of these modules.

These modules are excluded from the main theoretical discussion for two reasons: some are intuitive without formal mathematical backing, while others are adopted in existing systems and not novel enough. Thus, we provide their implementation details and design considerations in Appendix~\ref{engineering}.

\section{Experiments}

\subsection{Experimental Setup}
We evaluate the effectiveness of our stability-aware prompt optimization framework and semantic stability metric through a series of experiments based on the proposed general-purpose multi-agent system, Promptor. This system integrates all key components—including prompt stability evaluation, guided refinement, and feedback-based optimization—and serves as the unified platform for all results reported in this section.

All experiments use \textbf{GPT-4o} as the language model backbone. Unless noted otherwise, all baselines also use GPT-4o, ensuring fair comparison under identical model and I/O settings.

Our evaluation covers three complementary aspects:

\textbf{Performance on General tasks}: We test on diverse tasks—math reasoning, data analysis, code generation, and machine learning—and compare against existing systems such as AutoGen \cite{c6}, Data Interpreter (DI) \cite{c5}, EvoMAC \cite{c7}, and DA-Agent \cite{c9}.

\textbf{Performance on Domain-specific tasks}: We assess performance on expert-level tasks in biology, finance, and chemistry, where we compare against domain-specific systems with handcrafted prompts or fine-tuned models.

\textbf{Ablation studies}: We disable individual modules (e.g., subtask optimizer, stability evaluator, prompt reviewer, plan updater) to quantify their contributions to task success and stability.

Unlike baselines with complex coordination or handcrafted pipelines, our system uses a simple architecture focused on automated prompt design. Its strong performance highlights the practical value of prompt stability as a core optimization target.

\subsection{Validating semantic stability as a prompt evaluation metric}

To justify the use of semantic stability as a proxy for output stability, we conduct both qualitative and quantitative evaluations that demonstrate its alignment with intuitive notions of prompt consistency and task success.

\begin{figure}[h]
    \centering
    \includegraphics[width=\linewidth]{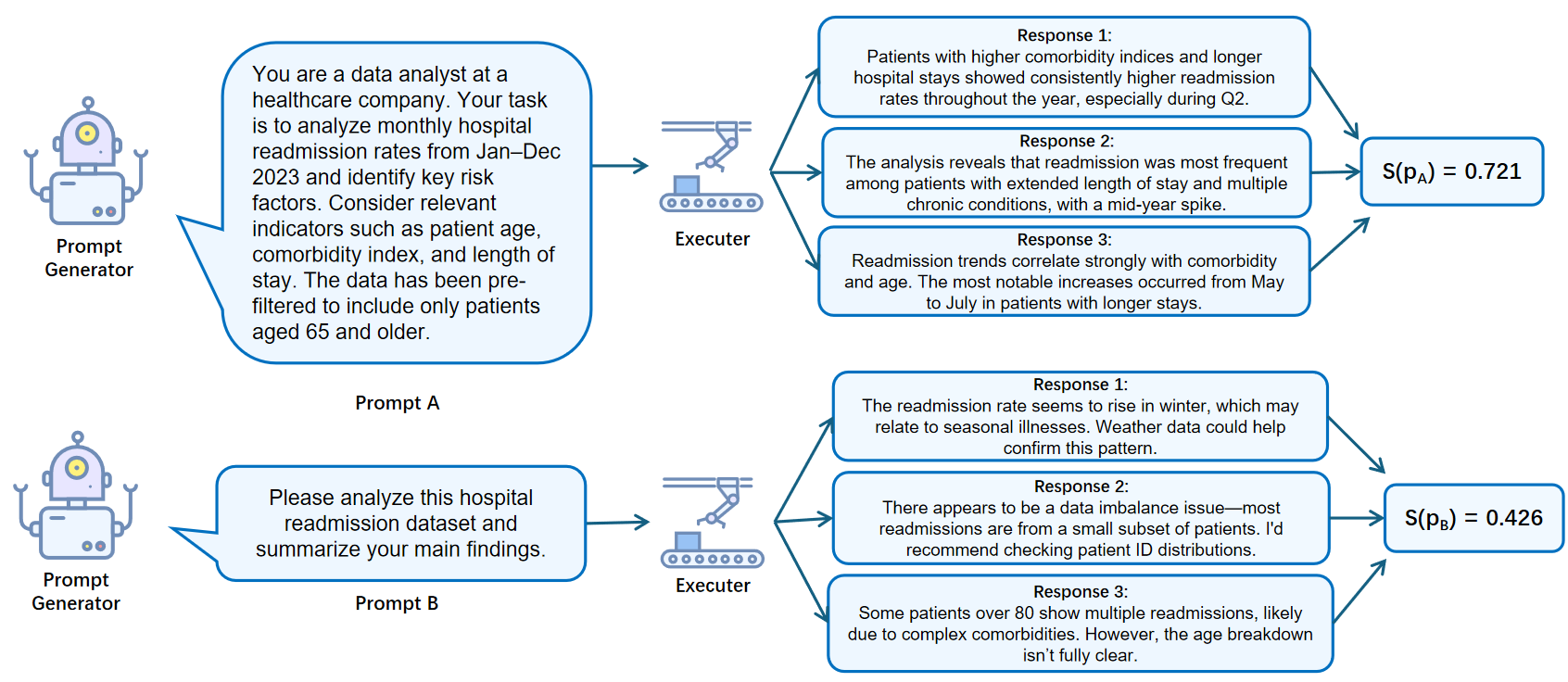}
    \caption{Illustration of semantic stability as a metric.}
    \label{fig:prompt_stability_example}
\end{figure}

\textbf{Qualitative Analysis.}
To evaluate semantic stability, we compare two prompts for the same data analysis task with different structures. As shown in Fig.~\ref{fig:prompt_stability_example}, Prompt B lacks clear role, scope, and domain context, leading to divergent outputs. Prompt A follows our structured format—with explicit role, requirements, knowledge, and history—and yields consistent results aligned with the planner’s intent. Their semantic stability score difference highlights that stability reliably reflects output variance caused by prompt ambiguity. This demonstrates that well-structured prompts improve both consistency and interpretability, making semantic stability an effective evaluation metric.

\textbf{Quantitative Analysis -- Correlation with Task Success Rate.}
We quantitatively assess semantic stability by computing its correlation with task success across 2000 prompts from our pipeline. For each prompt, we measure its stability score and average success rate in 6 executions. The results show a strong positive correlation ($r = 0.73$): more stable prompts are significantly more likely to succeed. To further evaluate the interpretability of our proposed metric to prompt performance, we compared \textit{Stability Score} with several traditional metrics, including mutual information~\cite{sorensen2022mintip}, prompt entropy~\cite{lu-etal-2022-fantastically}, clarity, specificity, and coherence~\cite{ghost2025promptqual}, by jointly using them as predictors in XGBoost regressor to predict the success rate. As shown in Table~\ref{tab:feature_importance}, 
\textit{Stability Score} exhibits the highest importance among all considered metrics, indicating that it provides the strongest explanatory signal compared to existing alternatives. This confirms that semantic stability reflects not just surface-level variance, but deeper ambiguities that affect performance—making it a reliable and interpretable metric for prompt refinement.

\begin{table}[ht]
  \centering
  \footnotesize
  \begin{tabular}{lc}
    \toprule
    Feature & Importance (mean $\pm$ std) \\
    \midrule
    Stability Score (ours) & 0.544 $\pm$ 0.062 \\
    Mutual Information     & 0.391 $\pm$ 0.027 \\
    Prompt Entropy         & 0.303 $\pm$ 0.012 \\
    Clarity                & 0.149 $\pm$ 0.021 \\
    Specificity            & 0.135 $\pm$ 0.019 \\
    Coherence              & 0.115 $\pm$ 0.011 \\
    \bottomrule
  \end{tabular}
  \caption{Feature importance of different metrics on XGB Regressor.}
  \label{tab:feature_importance}
\end{table}

\subsection{Effectiveness of Stability-Aware Prompt Optimization Framework}

\subsubsection{General Ability Assessment Result}
To evaluate our system’s ability to handle general tasks, we compared our system, Promptor, with several general-purpose or multi-functional systems across four domains: mathematical reasoning, data analysis, code writing, and basic machine learning. Scores are reported in Table~\ref{tab:general_ability}.

\begin{table}[htbp]
    \centering
    \footnotesize
    \begin{tabular}{lccccc}
        \toprule
        System & Math Reasoning & Data Analysis & Code Writing & Machine Learning & Avg \\
        \midrule
        Promptor & \textbf{0.63} & 0.88 & 0.94 & \textbf{0.86} & \textbf{0.84} \\
        DI       & \textbf{0.63} & \textbf{0.95} & 0.84 & 0.80 & 0.82 \\
        AutoGen  & 0.50 & 0.71 & 0.88 & 0.82 & 0.73 \\
        DA-Agent & 0.43 & 0.65 & 0.73 & 0.58 & 0.62 \\
        EvoMac   & 0.60 & 0.43 & \textbf{0.95} & 0.44 & 0.63 \\
        \bottomrule
    \end{tabular}
    \caption{Comparison of different systems across four general task domains.}
    \label{tab:general_ability}
\end{table}

\textbf{Math Reasoning} \quad  
We evaluate level-5 problems from four MATH subdomains \cite{c10}, which require multi-step symbolic reasoning beyond simple retrieval. Promptor matches DI—the strongest baseline—and outperforms the others. This highlights its ability to support structured problem-solving with automatically generated prompts. See Appendix Figure~\ref{fig:MATH}.

\textbf{Data Analysis} \quad  
On the InfiAgent-DABench benchmark \cite{c11}, which covers over 250 real-world CSV-based analysis tasks, Promptor reaches a high similarity score of 0.88. This indicates strong performance in tasks involving coding, debugging, and dynamic reasoning. The result highlights how stability-guided prompt refinement improves reliability in open-ended, data-driven scenarios.

\textbf{Code Writing} \quad  
On HumanEval \cite{c8}, a benchmark for functional code generation, Promptor reaches 0.94, rivaling EvoMac—a system fine-tuned for programming. Our result shows that prompt stability alone enables complex generation and self-correction.

\textbf{Machine Learning} \quad  
We test on ML-Benchmark, covering 10 tasks involving prediction, evaluation, and visualization. To ensure fairness, we remove restrictive instructions and retain only the core task descriptions. Each task is repeated five times, and we report the execution success rate (ESR) and prediction accuracy in Appendix Figure~\ref{fig:ml_benchmark}. Promptor achieves the highest score (0.86), demonstrating strong adaptability in complex ML scenarios.

\subsubsection{Professional Ability Assessment Result}

General-purpose systems often underperform task-oriented systems in specialized domains requiring structured and professional domain knowledge. To test whether our stability-guided framework overcomes this limitation, we evaluate Promptor on expert-level tasks in biology, chemistry, and finance, against both general-purpose baselines and task-oriented systems that use domain-specific pretraining or handcrafted prompts. Results are shown in Table~\ref{tab:professional}.

\textbf{Biology}  
We evaluate Promptor on single-cell RNA sequencing (scRNA-seq) analysis using the CellAgent benchmark, which includes 50+ datasets across diverse tissues and cell types \cite{c12}. Promptor achieves 86\% accuracy—outperforming general-purpose baselines and approaching CellAgent (92\%), despite not using domain-specific planning. This suggests that stability-aware prompting can support complex biomedical tasks by reducing ambiguity.

\textbf{Chemistry}  
On the SMILES-to-molecular-formula (S2MF) task from ChemLLMBench \cite{c14}, Promptor achieves the highest accuracy, surpassing even the fine-tuned ChemDFM model. Although differences in model size and pretraining scale may account for part of the gap, the results show that prompt stability substantially benefits symbolic reasoning even without domain-specific pretraining.

\textbf{Finance}  
We assess financial decision-making using Apple (AAPL) stock data and measure Annual Rate of Return (ARR). Promptor achieves +16.2\%, outperforming general-purpose systems and approaching FinAgent, a domain-optimized system \cite{c13}. Several baselines fail to complete runs, suggesting that stability-aware optimization enhances robustness in high-stakes environments.

Across domains, Promptor consistently narrows the gap with task-specific systems and exceeds general-purpose baselines, indicating that prompt stability helps general agents adapt to professional tasks without requiring domain-specific finetuning.

\begin{table*}[t]
\centering
\begin{minipage}{0.48\linewidth}
    \centering
    \footnotesize
    \begin{tabular}{lccc}
        \toprule
        System & Biology & Chemistry & Finance  \\
        \midrule
        Promptor & \textbf{0.86} & \textbf{0.75}  & \textbf{+16.2}  \\
        DI & 0.63 & 0.69 & -2.29 \\
        AutoGen & 0.68 & 0.60 & +8.74 \\
        DA-Agent & -- & 0.41 & --  \\
        EvoMac & 0.43 & 0.57 & -- \\
        \midrule
        Task-Oriented & 0.92 & 0.73 & +31.9   \\
        \bottomrule
    \end{tabular}
    \caption{Comparison across three professional domains.}
    \label{tab:professional}
\end{minipage}
\hfill
\begin{minipage}{0.48\linewidth}
    \centering
    \footnotesize
    \begin{tabular}{lcccc}
        \toprule
        Condition & \multicolumn{2}{c}{ESR \%} & \multicolumn{2}{c}{CR \%} \\
        \midrule
        Promptor(Our system) & 98 & $\rightarrow$ & 91 & $\rightarrow$ \\
        w/o Subtask Optimizer & 92 & $6\downarrow$ & 78 & $13\downarrow$ \\
        w/o Prompt Reviewer & 97 & $1\downarrow$ & 77 & $14\downarrow$ \\
        w/o S(p), w/ KL(p) & 97 & $1\downarrow$ & 83 & $8\downarrow$ \\
        w/o Plan Updater & 73 & $25\downarrow$ & 60 & $31\downarrow$ \\
        \bottomrule
    \end{tabular}
    \caption{Ablation experiment results for different disabled modules.}
    \label{tab:ablation_results}
\end{minipage}
\end{table*}

\subsubsection{Ablation Experiment Result}
We evaluate the contribution of each stability-aware module via ablation studies. Specifically, we remove the Subtask Optimizer, Prompt Reviewer, or Plan Updater individually, and assess the impact on system performance. We also test a variant that replaces semantic stability $S(p)$ with token-level KL divergence as the refinement criterion.

We use 50 multi-step code generation tasks from the HumanEval dataset. Each task is repeated five times, and average execution success rate (ESR) and the correctness rate (CR) are reported in Table~\ref{tab:ablation_results}.

The results show that each module in our framework plays a critical role—removing any one degrades performance. Furthermore, replacing semantic stability with token-level KL also hurts results, confirming $S(p)$ as a more effective refinement signal.

Importantly, while the removal of engineering modules (e.g., the Plan Updater) results in larger performance drops, this primarily reflects their role as essential infrastructure that enables the system to function end-to-end. 
In contrast, our stability-aware optimization strategy contributes at a different level: it improves the reliability and interpretability of prompt refinement once a working pipeline is in place. 
Thus, the apparent dominance of engineering components in ablation studies does not diminish the methodological value of stability-aware design; rather, the two are complementary, with stability providing principled gains beyond what engineering alone can achieve.

\section{Conclusion}

This paper introduces prompt stability as a core design principle for general-purpose multi-agent systems. Through theoretical analysis and empirical validation, we show that semantic stability provides a practical and interpretable proxy for measuring output variance and guiding prompt optimization. Our proposed system, Promptor, integrates stability-aware refinement and feedback-driven planning to improve execution consistency and task success. Experiments across both general and domain-specific tasks demonstrate that enhancing prompt stability can significantly improve system reliability, even in complex, high-stakes environments. These findings suggest that prompt stability offers a robust foundation for scaling multi-agent coordination in LLM-based systems.

\section*{Acknowledgment}
This work was partially supported by the National Artificial Intelligence Research Resource (NAIRR) Pilot under awards NAIRR250400 and NAIRR240283, and Standing Up to POTS.

\bibliography{reference}


\appendix
\section{Appendix}

\subsection{Proof of the bounded probability} \label{proof}
Let $Z_i = u_i v_i (\x_i - \hat{\x_i})$, then we have $\text{Var}(Z_i) = (u_i v_i)^2 \text{Var}(x_i)$. The deviation becomes:
\[
\left| \bs - \hat{\bs} \right| = \left| \sum_{i=1}^n Z_i \right|.
\]
The concentration inequality for the sum of independent centered variables \cite{c4}:
\[
\bP\left( \left| \sum_{i=1}^n Z_i \right| \geq \epsilon \right) \leq 2 \exp\left( \frac{-\epsilon^2}{2 \sum_{i=1}^n \text{Var}(Z_i)} \right),
\]
By applying this inequality, we can derive a bound on the probability that this deviation exceeds a threshold $\epsilon$:
\[
\bP\left( \left| \bs - \hat{\bs} \right| \geq \epsilon \right) \leq 2 \exp\left( \frac{-\epsilon^2}{2 \sum_{i=1}^n (u_i v_i)^2 \text{Var}(\x_i) } \right).
\]
\subsection{Engineering Specifications} \label{engineering}

\paragraph{\textbf{I/O Stream Specification.}} 
We enforce explicit input-output specifications for each subtask. By standardizing the format and content of outputs, we constrain the support of the output distribution to a subset $\mathcal{C}$, i.e., $\x_i \sim \Lambda^{(i)}(p_i \mid \x_i \in \mathcal{C})$. This restriction reduces semantic variance and improves stability across executions.

\paragraph{\textbf{Domain Knowledge Generation.}}  
We elicit domain-specific knowledge directly from the LLM using a Domain Knowledge Generator, which prompts the model to recall relevant facts from its internal corpus. Unlike retrieval-augmented generation (RAG) methods, our approach is independent of external databases and thus avoids issues of limited coverage or retrieval imprecision. This enables scalable, on-demand access to expert-level information, particularly in under-documented domains.

\paragraph{\textbf{Semi-Automatic Interaction Generation.}}  
Dynamically generating roles and their interactions remains a key challenge in general-purpose systems. Fully automated pipelines often lead to over-generated roles and unstable information flows. To address this, we adopt a semi-automatic approach: tasks are categorized into a small number of templates, each associated with a fixed interaction structure that supports both linear and nonlinear flows. The system selects a suitable template based on task type and instantiates roles accordingly. This approach reduces coordination errors while preserving flexibility. Each role also supports modular subfunctions that can be invoked on demand, further reducing redundancy and stabilizing agent collaboration.

\paragraph{\textbf{Requirement Augmentation.}}  
To mimic the iterative nature of human prompt refinement, our system incorporates a feedback loop that monitors execution outcomes. When a subtask fails or yields suboptimal results, the Reviewer Agent evaluates whether the failure stems from missing constraints. If so, new requirements are appended to the prompt, steering the model toward more reliable behavior. This process continues until performance reaches an acceptable threshold, preventing recurrence of similar issues.

\paragraph{\textbf{Context and History Management.}}  
Maintaining coherent information flow across subtasks is critical in long-horizon tasks. The Subtask Optimizer defines explicit data paths and formats during planning, enabling the Prompt Generator to structure prompts accordingly. After each execution, results are stored in a structured format and passed through a Summarizer Agent, which distills lengthy outputs into concise summaries. These summaries serve as compact memory units, enhancing context retention and reducing semantic drift in downstream subtasks.

\paragraph{\textbf{User Interface Design.}}

To demonstrate the practical applicability of our framework, we implemented a web-based user interface named \textit{MyDataPilot}. 
The system follows a three-tier architecture with a React frontend, Node.js middleware, and a Python backend. 
The UI is designed to support stability-aware multi-agent task decomposition, ensuring consistent and reliable responses through semantic stability optimization. 
Key functionalities include:
\begin{itemize}
    \item Intelligent task decomposition with stability-aware planning.
    \item Real-time conversational interaction with agents.
    \item Task execution management and progress tracking.
    \item Automated exploratory data analysis (EDA) with visualization.
    \item User authentication and session management.
    \item Code generation with file saving and download support.
\end{itemize}

This system has been deployed as a complete web application, bridging the gap between research and practice. 
An interactive demonstration is available at \url{https://xucheng63.github.io/MyDataPilot/}.

\subsection{Detailed Results on Some Benchmarks}

\begin{figure}[ht]
    \centering
    \includegraphics[width=0.8\linewidth]{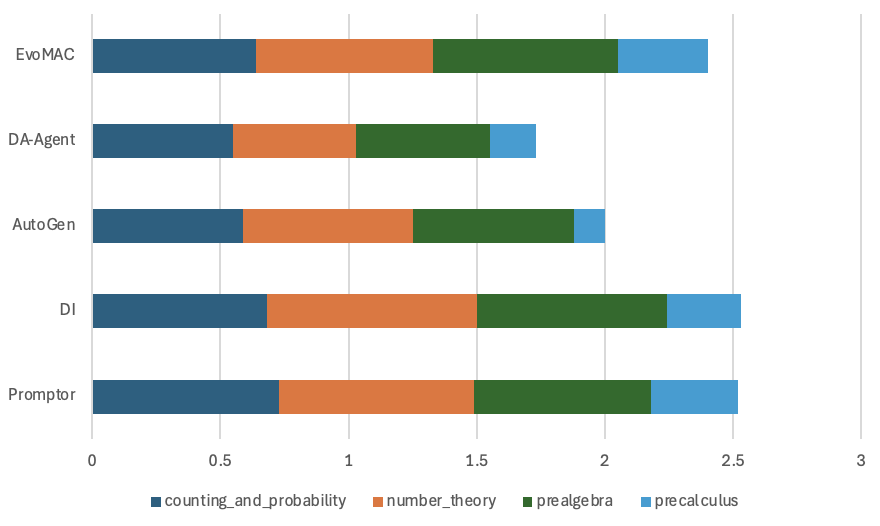}
    \caption{
Performance on the MATH benchmark.
The horizontal axis shows cumulative accuracy defined as 
$\mathrm{CumAcc} = \sum_{k=1}^{4} \mathrm{Acc}_k$, 
where $\mathrm{Acc}_k \in [0,1]$ denotes the accuracy on each of the four subtasks 
(counting and probability, number theory, pre-algebra, and pre-calculus). 
Each colored segment corresponds to one subtask.
}
    \label{fig:MATH}
\end{figure}

The MATH benchmark consists of four sub-tasks, covering algebra, geometry, number theory, and probability.
Figure \ref{fig:MATH} shows the performance of different systems on each sub-task individually, as well as the aggregated overall score. The comparison allows us to observe how each system behaves across tasks with varying reasoning difficulty and mathematical structure.

For the ML-Bench benchmark, we evaluate two important dimensions of system capability:

Execution Success Rate (ESR) – whether the system can successfully complete the task.

Accuracy – whether the generated output is correct when execution is successful.

\begin{figure}[ht]
    \centering
    \includegraphics[width=0.8\linewidth]{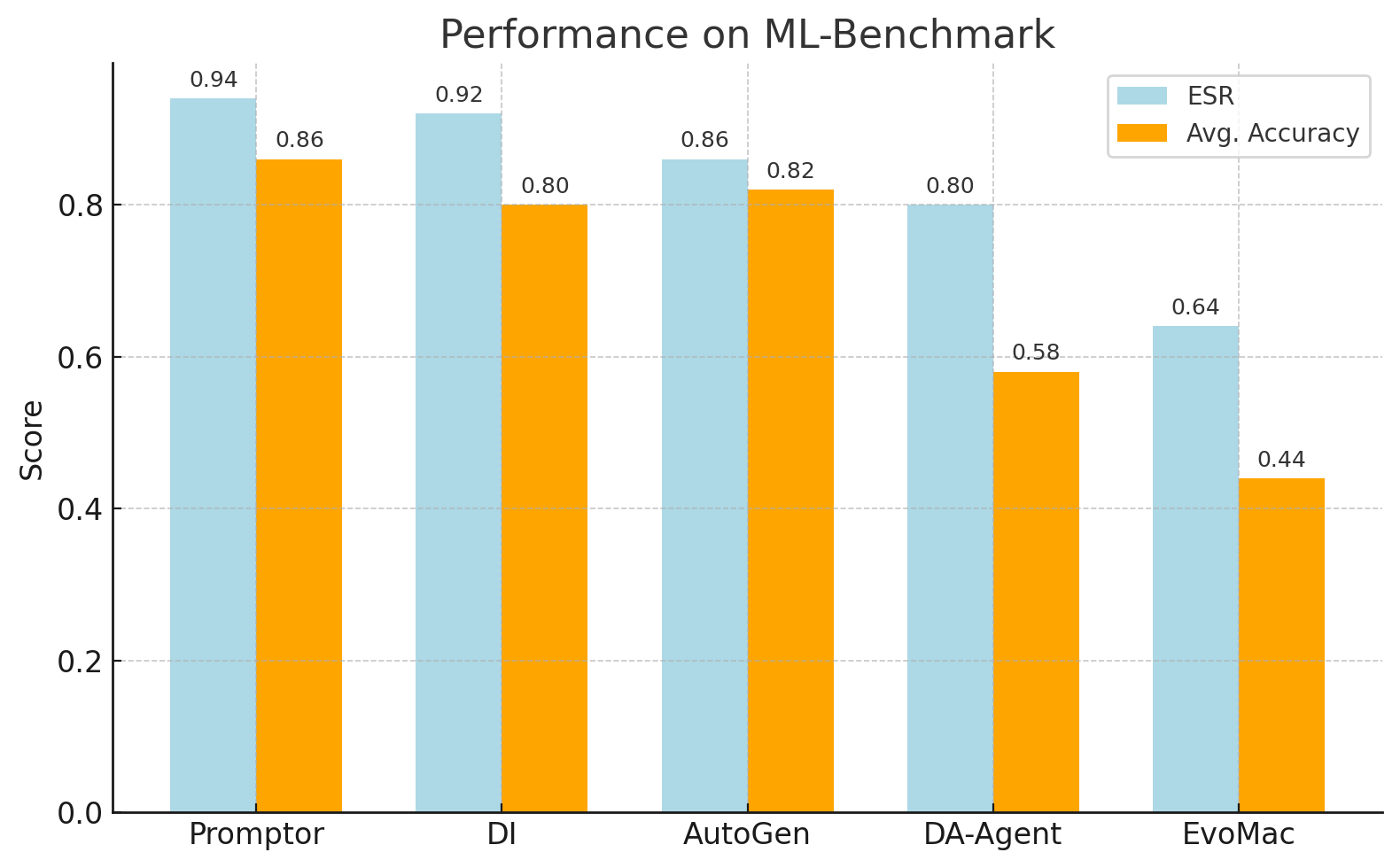}
    \caption{Performance on ML-Bench benchmark.}
    \label{fig:ml_benchmark}
\end{figure}

Figure \ref{fig:ml_benchmark} presents the ESR and accuracy of five systems, enabling a clear comparison across both reliability and correctness.

\end{document}